\documentclass[conference]{IEEEtran}
\IEEEoverridecommandlockouts
\usepackage{cite}
\usepackage{amsmath,amssymb,amsfonts}
\usepackage{algorithmic}
\usepackage{graphicx}
\usepackage{textcomp}
\usepackage{xcolor}
\usepackage{makecell}
\usepackage{multirow}
\usepackage{color}
\usepackage[table]{xcolor}
\usepackage{subcaption}
\usepackage{placeins}

\usepackage{subcaption}
\def\BibTeX{{\rm B\kern-.05em{\sc i\kern-.025em b}\kern-.08em
    T\kern-.1667em\lower.7ex\hbox{E}\kern-.125emX}}

\usepackage{xcolor}
\usepackage{pifont}
\usepackage[table]{xcolor}
\newcommand{\cmark}{\textcolor{green!60!black}{\textbf{\checkmark}}}
\newcommand{\xmark}{\textcolor{red}{\textbf{\texttimes}}}

\begin{document}

%\title{Learning Surface Materials from Random Sparse Visual Observations: A Comparative Study of ConvAE, MAE, ViT, and Swin Transformers\\

\title{Seeing Through Extreme Visual Sparsity: Surface Understanding from a Single Random Visual Patch\\

% \author{Anonymous Authors}

% \author{
% \IEEEauthorblockN{Sindhuja Penchala}
% \IEEEauthorblockA{
% \textit{Department of Computer Science} \\
% \textit{The University of Alabama}\\
% Tuscaloosa, Alabama, USA \\
% spenchala@crimson.ua.edu
% }
% \and
% \IEEEauthorblockN{Sudip Mittal}
% \IEEEauthorblockA{
% \textit{Department of Computer Science} \\
% \textit{The University of Alabama}\\
% Tuscaloosa, Alabama, USA \\
% sudip.mittal@ua.edu
% }
% \and
% \IEEEauthorblockN{Noorbakhsh Amiri Golilarz}
% \IEEEauthorblockA{
% \textit{Department of Computer Science} \\
% \textit{The University of Alabama}\\
% Tuscaloosa, Alabama, USA \\
% namirigolilarz@ua.edu
% }
% }

\author{
\IEEEauthorblockN{
Sindhuja Penchala\textsuperscript{1},
Sudip Mittal\textsuperscript{1},
and Noorbakhsh Amiri Golilarz\textsuperscript{1}
}
\IEEEauthorblockA{
\textsuperscript{1}Department of Computer Science,
The University of Alabama, Tuscaloosa, AL, USA \\
spenchala@crimson.ua.edu,
\{sudip.mittal, namirigolilarz\}@ua.edu
}
}

}

% \author{\IEEEauthorblockN{1\textsuperscript{st} Sindhuja Penchala}
% \and
% \and
% \and
% \and
% \and

\maketitle

\begin{abstract}
% Surface material recognition from incomplete visual observations is a challenging problem in robotic perception and environmental understanding. This paper presents a comparative study of four pretrained architectures-Convolutional Autoencoder (ConvAE), Vision Transformer (ViT), Swin Transformer, and Masked Autoencoder (MAE) for simultaneous surface material reconstruction and classification under extreme visual sparsity. Experiments were conducted on the Touch-and-Go dataset using a sparse observation protocol in which only 10\% of the original image remained visible while the remaining regions were masked. The models were evaluated using reconstruction quality, classification performance, model complexity, and inference efficiency metrics. Experimental results show that transformer-based models outperform conventional convolutional architectures in material classification. Swin Transformer achieved the highest classification accuracy of 89.55\% and F1-score of 0.8955, while MAE and ViT produced superior reconstruction quality. Furthermore, all evaluated models demonstrated real-time inference capability, with processing times below 5 ms per image. The results highlight the effectiveness of pretrained transformer representations for extracting meaningful material information from highly sparse visual inputs and provide insights into the trade-offs between reconstruction fidelity, classification accuracy, and computational efficiency.

Surface material recognition from incomplete visual observations remains a challenging problem in robotic perception and environmental understanding. This paper discusses Sparse Surface Understanding Framework (SSUF), a unified dual-task learning framework that adapts four pretrained architectures-Convolutional Autoencoder (ConvAE), Vision Transformer (ViT), Swin Transformer, and Masked Autoencoder (MAE) for simultaneous surface reconstruction and material classification. Experiments were conducted on the Touch-and-Go dataset using a sparse observation protocol in which only 10\% of the original image remained visible while the remaining regions were masked. To enable a fair comparison, reconstruction-oriented models were extended with classification heads, whereas classification-oriented models were augmented with reconstruction decoders. The resulting architectures were assessed using reconstruction quality, classification performance, model complexity, and inference efficiency metrics. Experimental results revealed distinct strengths across the models. Swin Transformer achieved the best classification performance with an accuracy of 89.21\%, an F1-score of 0.8922, and a ROC-AUC of 0.9813. In contrast, MAE produced the highest reconstruction scores among evaluated models, with a PSNR of 16.06 dB and an SSIM of 0.4501, while ViT provided the best overall balance between reconstruction and classification performance. Furthermore, all models achieved real-time inference, requiring less than 5 ms per image. Overall, the results show that pretrained architectures can support material recognition under severe visual sparsity, while accurate image reconstruction remains challenging.
% These findings demonstrate the effectiveness of pretrained transformer representations for sparse surface understanding and highlight the trade-offs between reconstruction fidelity, recognition accuracy, and computational efficiency.

\end{abstract}


\section{Introduction}

Surface material recognition plays an important role in robotics, autonomous navigation, human-computer interaction, and embodied artificial intelligence. Accurate identification of materials such as concrete, grass, wood, and rock enables intelligent systems to reason about terrain properties, estimate physical interactions, and improve decision-making in real-world environments. Recent advances in computer vision have significantly improved visual perception through deep learning, particularly with the introduction of large-scale pretrained models that learn rich visual representations from extensive image corpora \cite{dosovitskiy2020image, liu2022swin, he2022masked}.

Traditional surface material recognition methods generally assume access to complete visual observations, enabling models to exploit rich texture, color, and structural cues for accurate classification. In practical environments, however, surfaces are frequently observed under incomplete conditions caused by occlusions, limited sensor coverage, motion blur, adverse illumination, or environmental interference. As a result, intelligent systems must often reason from highly restricted visual evidence, where substantial portions of the scene are unavailable. The ability to make reliable decisions from incomplete observations is also recognized as an important requirement for more autonomous and adaptive intelligent systems \cite{golilarz2025bridging}. Under such conditions, reliable surface understanding requires not only recognition of material characteristics but also the ability to infer missing visual information from the available observations.

% Recent studies have shown that meaningful surface understanding can be achieved even from highly limited visual observations, highlighting the potential of sparse visual cues for both reconstruction and recognition tasks \cite{penchala2026one}.
% Transformer-based surface understanding frameworks, including SurFormer V1 and SurFormer V2, have demonstrated the effectiveness of attention mechanisms and multimodal feature fusion for material recognition \cite{kansana2025surformerv1,kansana2025surformer}. More broadly, pretrained architectures such as Vision Transformer (ViT), Swin Transformer, and Masked Autoencoder (MAE) have achieved remarkable success across a wide range of computer vision tasks, including image recognition, representation learning, and image reconstruction \cite{dosovitskiy2020image, liu2022swin, he2022masked}. Convolutional architectures based on ResNet also remain effective for extracting discriminative texture and appearance features \cite{he2016deep, koonce2021resnet}. Despite their success, these models have largely been evaluated within their original objectives and under settings where sufficient visual information is available.

Recent studies have shown that surface reconstruction and recognition are possible even from highly sparse visual observations \cite{penchala2026one}. Our prior work, \textit{One Patch Is All You Need}, introduced SMARC, a Partial Convolutional U-Net that jointly reconstructs surface appearance and predicts material categories from a fixed central patch containing only 10\% of the image. While that study established the feasibility of dual-task learning under extreme visual sparsity, it was limited to a fixed patch location and comparatively smaller baseline architectures. In contrast, the present work uses a randomly positioned continuous 10\% patch and introduces SSUF to systematically adapt and compare full-scale pretrained ConvAE, ViT, Swin Transformer, and MAE architectures within the same dual-task setting. This extension evaluates whether sparse surface understanding remains effective when the visible region is not spatially fixed, while also comparing reconstruction, classification, model complexity, and inference efficiency. Transformer-based surface understanding frameworks such as SurFormer V1 and V2 have further demonstrated the effectiveness of attention and multimodal feature fusion for material recognition \cite{kansana2025surformerv1,kansana2025surformer}. More broadly, ViT, Swin Transformer, MAE, and ResNet-based architectures have shown strong performance in visual representation learning, reconstruction, and recognition \cite{dosovitskiy2020image,liu2022swin,he2022masked,he2016deep,koonce2021resnet}.

Although these pretrained models have achieved the best performance across various computer vision tasks, several limitations remain. First, most existing studies evaluate reconstruction and classification independently rather than within a unified dual-task learning framework. Second, current masked-image modeling approaches typically assume randomly distributed visible patches during training, whereas real-world observations often consist of localized continuous regions. Third, limited research has investigated how different pretrained architectures behave under extreme visual sparsity for surface material understanding. In particular, a comprehensive comparison of convolutional and transformer-based pretrained models under identical sparse-observation settings is largely absent from the literature.

To address these limitations, this work investigates surface material reconstruction and classification using only a randomly selected continuous patch corresponding to approximately 10\% of the original image area. We introduce a Sparse Surface Understanding Framework (SSUF), a unified dual-task sparse surface understanding framework built upon four pretrained architectures: a ResNet50-based Convolutional Autoencoder (ConvAE), Vision Transformer (ViT), Swin Transformer, and Masked Autoencoder (MAE). Since these architectures were originally designed for either reconstruction or classification, they were adapted into a common dual-task framework through complementary task-specific extensions. Specifically, MLP-based classification heads were added to ConvAE and MAE, while CNN-based reconstruction decoders were integrated into ViT and Swin Transformer. Experiments were conducted on the Touch-and-Go dataset \cite{yang2022touch}, which contains real-world surface material images collected under natural environmental conditions. Unlike prior studies that focus exclusively on reconstruction or classification, the framework enables a unified evaluation of both tasks while also considering model complexity and inference efficiency. The main contributions of this work are summarized as follows:

\begin{itemize}
\item Development of a unified Sparse Surface Understanding Framework (SSUF) for joint surface reconstruction and material classification under extreme visual sparsity.

\item Transformation of reconstruction-oriented and classification-oriented pretrained models into a common dual-task learning paradigm through complementary architectural extensions.

\item Introduction of a continuous-patch sparse observation protocol that preserves only 10\% of the original visual information.

\item Comprehensive benchmarking of ConvAE, ViT, Swin Transformer, and MAE in terms of reconstruction fidelity, recognition accuracy, computational complexity, and inference efficiency.
\end{itemize}

% \begin{itemize}
%     \item To investigate the effectiveness of pretrained visual representations for reconstructing surface images from highly sparse observations.
%     \item To evaluate the ability of pretrained models to classify surface materials using only 10\% visible visual information.
%     \item To compare reconstruction quality, classification performance, computational complexity, and inference efficiency across convolutional and transformer-based architectures.
%     \item To identify the most suitable pretrained model for joint reconstruction and classification under extreme visual sparsity.
% \end{itemize}

% Based on the architectural characteristics of the evaluated models, the following hypotheses are investigated:

% \begin{itemize}
%     \item \textbf Transformer-based models will outperform convolutional architectures in surface material classification due to their ability to capture long-range contextual relationships.
%     \item \textbf MAE will achieve superior reconstruction performance because its self-supervised pretraining objective explicitly focuses on recovering missing visual content.
%     \item \textbf Swin Transformer will provide the best balance between classification performance and computational efficiency owing to its hierarchical representation learning strategy.
%     \item \textbf Joint reconstruction and classification learning will enable robust surface understanding even when only 10\% of the image is visible.
% \end{itemize}

The remainder of this paper is organized as follows. Section II reviews related work on surface material recognition, image reconstruction, and pretrained vision models. Section III describes sparse observation protocol, experimental framework and training strategy. Section IV presents the Touch-and-Go dataset, comparative evaluation results and performance analysis. Finally, Section V concludes the paper and outlines future research directions.

\section{Related Work}

\subsection{Surface Material Recognition}

Surface material recognition aims to identify physical surface properties from visual observations. Early deep learning approaches demonstrated that material-specific texture and appearance patterns can be effectively learned from large-scale datasets such as Materials in Context (MINC) \cite{bell2015material}. More recently, transformer-based and multimodal frameworks have improved recognition performance by leveraging visual and tactile information. SurFormer and SurFormer v2 demonstrated the effectiveness of attention mechanisms and vision-touch fusion for surface classification \cite{kansana2025surformerv1, kansana2025surformer}. However, these approaches primarily focus on classification and assume access to complete visual observations.

\subsection{Image Reconstruction from Sparse Observations}

Image reconstruction techniques aim to recover missing visual content from incomplete observations. Context Encoders introduced image completion through representation learning, while Partial Convolutions improved reconstruction quality by operating only on valid pixels \cite{pathak2016context, liu2018image}. More recent transformer-based methods have further enhanced reconstruction performance for large missing regions and complex visual structures \cite{yang2024review}. Despite these advances, reconstruction and recognition are typically studied as separate tasks.

\subsection{Pretrained Vision Models}

Pretrained vision models have become a fundamental component of modern computer vision due to their ability to transfer knowledge learned from large-scale datasets. ResNet-based architectures remain effective for capturing local texture and appearance features, while Vision Transformer (ViT) and Swin Transformer leverage self-attention mechanisms to model global and hierarchical visual relationships, respectively \cite{koonce2021resnet,dosovitskiy2020image,liu2022swin}. In addition, Masked Autoencoders (MAE) have demonstrated the effectiveness of self-supervised learning through masked image reconstruction, enabling robust representation learning from incomplete visual observations \cite{he2022masked}.

% \subsection{Pretrained Vision Models}

% Pretrained models have significantly advanced visual understanding through transfer learning. ResNet-based architectures remain effective for texture analysis and local feature extraction \cite{koonce2021resnet}, while Vision Transformer (ViT) and Swin Transformer achieve strong recognition performance through global and hierarchical self-attention mechanisms, respectively \cite{dosovitskiy2020image, liu2022swin}. Masked Autoencoders (MAE) further demonstrated the effectiveness of self-supervised reconstruction from incomplete observations \cite{he2022masked}. However, these architectures were originally developed for either reconstruction or classification, and their behavior within a unified dual-task framework remains largely unexplored. Furthermore, comparative studies evaluating convolutional and transformer-based pretrained models under extreme visual sparsity are scarce. To address this gap, this work adapts ConvAE, ViT, Swin Transformer, and MAE within the Sparse Surface Understanding Framework and systematically evaluates their reconstruction and classification capabilities using only 10\% visible image content.

\begin{table*}[ht]
\centering
\caption{Comparison of the evaluated architectures and the modifications introduced in this study.}
\label{tab:model_comparison}
\resizebox{\textwidth}{!}{%
\begin{tabular}{lcccc}
\hline
\rowcolor{gray!15}
\textbf{Component} &
\textbf{ConvAE} &
\textbf{ViT} &
\textbf{Swin} &
\textbf{MAE} \\
\hline

Primary Objective
& Reconstruction
& Classification
& Classification
& Self-Supervised Reconstruction \\

Encoder Type
& ResNet50
& Transformer
& Hierarchical Transformer
& Transformer \\

Patch Embedding
& \xmark
& \cmark
& \cmark
& \cmark \\

Attention Mechanism
& \xmark
& Global
& Shifted Window
& Global \\

Original Decoder
& \cmark
& \xmark
& \xmark
& \cmark \\

Original Classifier
& \xmark
& \cmark
& \cmark
& \xmark \\

Latent Representation
& Feature Maps
& Token Embeddings
& Hierarchical Features
& Latent Tokens \\

\hline
\rowcolor{gray!15}
\multicolumn{5}{c}{\textbf{Modifications Introduced in This Study}} \\
\hline

Added Module
&
\textbf{MLP Classifier}
&
\textbf{CNN Decoder}
&
\textbf{CNN Decoder}
&
\textbf{MLP Classifier}
\\

Dual-Task Learning
& \cmark
& \cmark
& \cmark
& \cmark
\\

\hline
\end{tabular}
}
\end{table*}

\section{Methodology}

\begin{figure*}[t]
\centering
\includegraphics[width=0.88\textwidth]{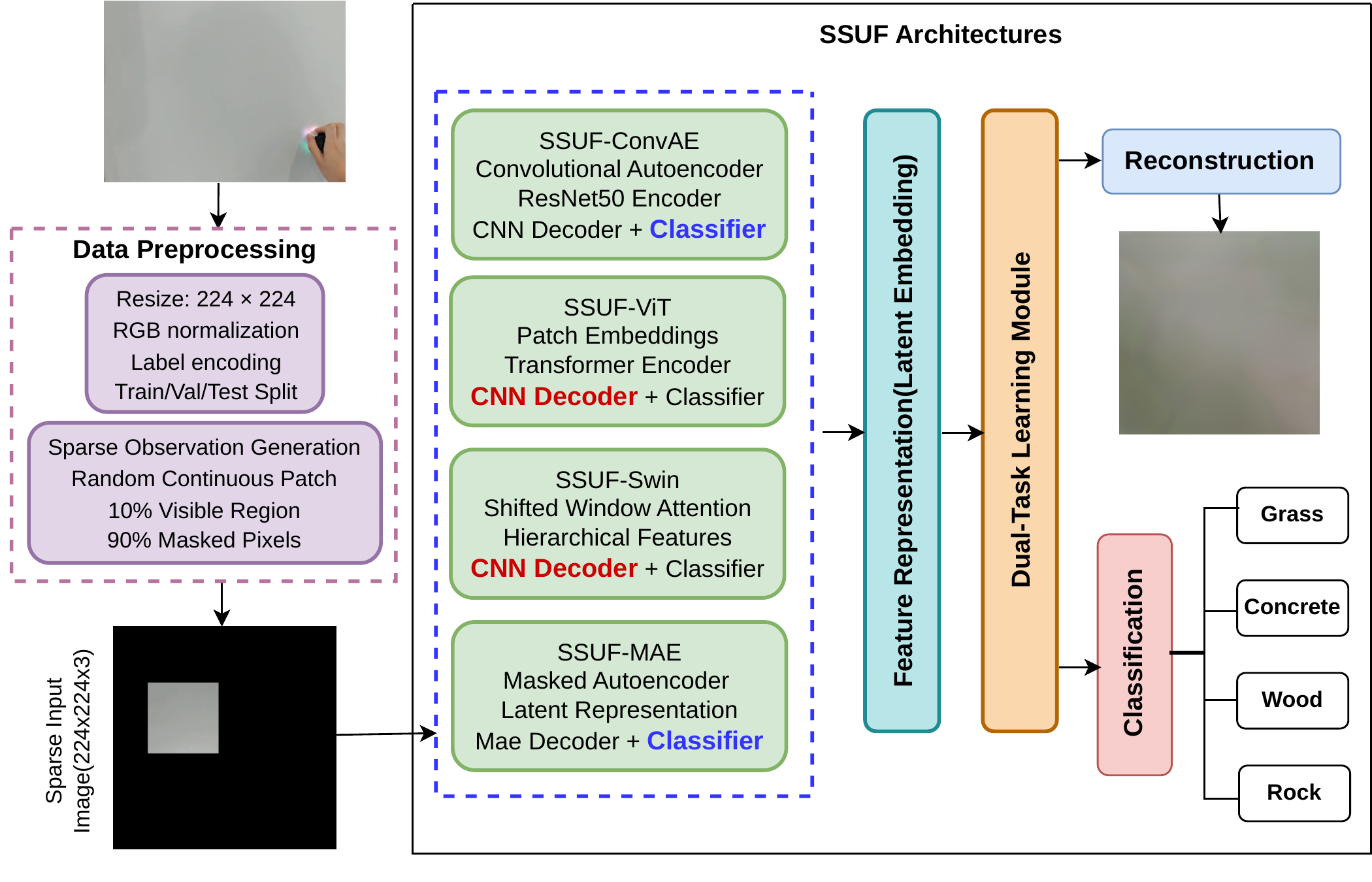}
\caption{Overview of the Sparse Surface Understanding Framework (SSUF) for simultaneous surface reconstruction and material classification from sparse visual observations. Images from the Touch-and-Go dataset are converted into sparse inputs by retaining only a random continuous patch corresponding to 10\% of the original image area. The resulting observations are processed by four adapted architectures: SSUF-ConvAE, SSUF-ViT, SSUF-Swin, and SSUF-MAE. Learned feature representations are jointly utilized for image reconstruction and classification of four surface material categories (Grass, Concrete, Wood, and Rock). Task-specific modules introduced to enable dual-task learning are highlighted in red (reconstruction decoders) and blue (classification heads).}
\label{fig:workflow}
\end{figure*}

The proposed experimental framework investigates the ability of pretrained vision models to reconstruct and recognize surface materials from highly limited visual observations. To ensure a consistent evaluation, all models were trained and tested using the same sparse-input generation procedure, dataset partitions, and optimization settings. The methodology consists of three main components: sparse observation generation, architecture-specific model design, and dual-task training. Fig. ~\ref{fig:workflow} illustrates the overall workflow of the framework.

% This study introduces a unified dual-task sparse surface understanding framework for simultaneous image reconstruction and surface material classification under extreme visual sparsity. Four pretrained architectures were investigated: a ResNet50-based Convolutional Autoencoder (ConvAE), Vision Transformer (ViT), Swin Transformer, and Masked Autoencoder (MAE). Since these architectures were originally designed for either reconstruction or classification, they were adapted through task-specific architectural extensions to perform both tasks simultaneously. Under the introduced framework, each model receives only a small visible image region and learns to reconstruct the original surface image while predicting the corresponding material category. Figure~\ref{fig:workflow} illustrates the overall workflow of the framework.

\subsection{Sparse Observation Generation and Model Adaptation}

To simulate extreme visual sparsity, only 10\% of each image was retained while the remaining 90\% was masked. A continuous square patch was randomly selected within the image boundaries, and pixels outside the selected region were set to zero. Let $I \in \mathbb{R}^{H \times W \times C}$ denote the original image and $M \in {0,1}^{H \times W}$ represent the binary visibility mask. The masked input image is computed as

\begin{equation}
I_{masked}=I \odot M
\end{equation}

where $\odot$ denotes element-wise multiplication. The same sparse observation protocol was applied to all four models to ensure a fair comparison.

For an image of size $H\times W$, the visible patch area is defined as $0.10HW$. A square patch with side length $s=\lceil\sqrt{0.10HW}\rceil$ is randomly selected within the image boundaries. For $224\times224$ inputs, this corresponds to a $71\times71$ patch, or approximately $10.05\%$ of the image area. The same random patch generation procedure was applied to all four architectures.

The evaluated architectures were not originally designed to perform simultaneous image reconstruction and material classification. ConvAE and MAE were primarily developed for image reconstruction, whereas ViT and Swin Transformer were designed for image classification. To enable a unified evaluation under the sparse-observation setting, all architectures were adapted into a common dual-task learning framework capable of performing both tasks simultaneously.

For the reconstruction-oriented architectures (ConvAE and MAE), an MLP-based classification head was attached to the latent feature representation to enable material recognition. Conversely, for the classification-oriented architectures (ViT and Swin Transformer), a CNN-based reconstruction decoder was integrated to recover the missing image regions from the learned feature representations. As a result, all models were trained to jointly reconstruct the original surface image and predict the corresponding material category from only 10\% visible image content. This unified design facilitates a comprehensive comparison of convolutional and transformer-based architectures in terms of reconstruction quality, classification performance, model complexity, and inference efficiency.

\subsection{SSUF-Convolutional Autoencoder Framework}

Within the SSUF framework, ConvAE serves as the convolution-based baseline. A pretrained ResNet50 encoder is employed to extract hierarchical feature representations from the sparse input image. These latent features are shared by two task-specific branches: a CNN-based decoder for image reconstruction and an MLP classification head for surface material recognition. The decoder progressively upsamples the latent representation through transposed convolution layers to recover the original image, while the classification branch applies global average pooling followed by fully connected layers and a Softmax classifier to predict the material category. Owing to their strong locality-preserving inductive biases, convolutional encoders are particularly effective for capturing texture and appearance characteristics that are important for surface material recognition \cite{koonce2021resnet}.

\subsection{SSUF-Vision Transformer Framework}

SSUF-ViT employs a pretrained Vision Transformer backbone initialized with ImageNet weights \cite{dosovitskiy2020image}. The sparse input image is represented as a sequence of patch embeddings and processed through transformer encoder layers to capture global contextual relationships via self-attention. To support the dual-task framework, a CNN-based decoder was integrated with the transformer encoder, enabling simultaneous image reconstruction and surface material classification.

\subsection{SSUF-Swin Transformer Framework}

SSUF-Swin utilizes a pretrained Swin Transformer backbone based on shifted-window self-attention \cite{liu2022swin}. Unlike ViT, Swin learns hierarchical multi-scale representations while maintaining computational efficiency. The extracted features are jointly used for reconstruction and classification, allowing the model to recover missing surface information while recognizing material categories from sparse observations.

\subsection{SSUF-Masked Autoencoder Framework}

SSUF-MAE is built upon the pretrained Masked Autoencoder architecture \cite{he2022masked}, which learns visual representations through masked image reconstruction. To enable dual-task learning, an MLP-based classification head was attached to the latent representation. The resulting framework simultaneously reconstructs the original surface image and predicts the corresponding material class from the sparse input.

Since each pretrained backbone has a different native architecture, task-specific modules were added only as required to enable the common dual-task reconstruction and classification setting.

\subsection{Training Strategy}

To ensure a fair comparison, all models were trained and evaluated using identical training, validation, and test partitions of the Touch-and-Go dataset. Input images were resized to $224 \times 224$ pixels and normalized according to the preprocessing requirements of the corresponding pretrained backbone. Sparse observations were generated by retaining only a random continuous patch corresponding to 10\% of the original image area while masking the remaining pixels.

A transfer learning strategy was adopted for all architectures. The pretrained ResNet50, ViT, Swin Transformer, and MAE backbones were initialized using publicly available ImageNet-pretrained weights, while task-specific modules were incorporated to enable simultaneous reconstruction and classification within the SSUF framework. As summarized in Table~\ref{tab:model_comparison}, MLP-based classification heads were added to ConvAE and MAE, whereas CNN-based reconstruction decoders were integrated into ViT and Swin Transformer to create a unified dual-task learning framework. Training was performed using the Adam optimizer. A dual-task objective function was employed to jointly optimize reconstruction and classification performance:

\begin{equation}
L_{total} = \lambda_{rec}L_{rec} + \lambda_{cls}L_{cls}
\end{equation}

where, $L_{rec}$ denotes the reconstruction loss computed using Mean Squared Error (MSE), $L_{cls}$ represents the categorical cross-entropy classification loss, and $\lambda_{rec}$ and $\lambda_{cls}$ control the contribution of each task. In all experiments, $\lambda_{rec}$ and $\lambda_{cls}$ were set to 1.0, giving equal importance to reconstruction and classification objectives. Class weights were incorporated during training to mitigate class imbalance among the four material categories. This training strategy enables all architectures to learn both structural image reconstruction and semantic material recognition from highly sparse visual observations.
While this study focuses on sparse surface understanding, the proposed SSUF framework can serve as a perception module for future autonomous robotic systems. Its ability to reason from limited visual observations may support integration with neurocognitive-inspired intelligence frameworks for higher-level reasoning and decision making under uncertainty \cite{golilarz2025bridging,golilarz2026towards}.

\section{Experimental Results and Discussions}

This section presents a comprehensive evaluation of the Sparse  SSUF under an extreme sparse-observation setting. The performance of four adapted architectures: SSUF-ConvAE, SSUF-ViT, SSUF-Swin, and SSUF-MAE is analyzed in terms of reconstruction quality, material classification performance, model complexity, and inference efficiency. 

\begin{table}[ht]
\centering
\caption{Reconstruction performance of the evaluated SSUF architectures
on the Touch-and-Go dataset using only 10\% visible image content.
Best results are shown in bold.}
\label{tab:reconstruction_results}

\footnotesize
\renewcommand{\arraystretch}{1.2}

\begin{tabular*}{\columnwidth}
{@{\extracolsep{\fill}}lcccc@{}}
\hline
\rule{0pt}{2.8ex}
\textbf{Model} &
\textbf{PSNR (dB) $\uparrow$} &
\textbf{SSIM $\uparrow$} &
\textbf{MAE $\downarrow$} &
\textbf{MSE $\downarrow$} \\
\hline
SSUF-ConvAE & 15.91 & 0.4380 & 0.13123 & 0.03056 \\
SSUF-ViT    & 16.03 & 0.4269 & \textbf{0.12861} & \textbf{0.02940} \\
SSUF-Swin   & 15.52 & 0.4255 & 0.13941 & 0.03307 \\
SSUF-MAE    & \textbf{16.06} & \textbf{0.4501} & 0.12924 & 0.03009 \\
\hline
\end{tabular*}

\end{table}

\begin{table}[ht]
\centering
\caption{Classification performance of the evaluated SSUF architectures
under the 10\% visible-patch setting. Precision, recall, and F1-score
are weighted averages. Best results are shown in bold.}
\label{tab:classification_results}

\footnotesize
\renewcommand{\arraystretch}{1.2}
\setlength{\tabcolsep}{7pt}

\begin{tabular}{lcccc}
\hline
\rule{0pt}{2.8ex}
\textbf{Model} &
\textbf{Acc. (\%) $\uparrow$} &
\textbf{Precision $\uparrow$} &
\textbf{Recall $\uparrow$} &
\textbf{F1 $\uparrow$} \\
\hline
SSUF-ConvAE & 83.90 & 0.8405 & 0.8390 & 0.8385 \\
SSUF-ViT    & 86.30 & 0.8640 & 0.8630 & 0.8630 \\
SSUF-Swin   & \textbf{89.21} & \textbf{0.8930} &
\textbf{0.8921} & \textbf{0.8922} \\
SSUF-MAE    & 76.71 & 0.7650 & 0.7671 & 0.7651 \\
\hline
\end{tabular}

\end{table}

% \begin{table}[ht]
% \centering
% \caption{Reconstruction performance of SSUF-ConvAE, SSUF-ViT, SSUF-Swin, and SSUF-MAE on the Touch-and-Go dataset using only 10\% visible image content. Best results are shown in bold.}
% \label{tab:reconstruction_results}
% \resizebox{\columnwidth}{!}{%
% \begin{tabular}{lcccc}
% \hline
% \textbf{Model} & \textbf{PSNR (dB) $\uparrow$} & \textbf{SSIM $\uparrow$} & \textbf{MAE $\downarrow$} & \textbf{MSE $\downarrow$} \\
% \hline
% SSUF-ConvAE & 15.91 & 0.4380 & 0.13123 & 0.03056 \\
% SSUF-ViT    & 16.03 & 0.4269 & \textbf{0.12861} & \textbf{0.02940} \\
% SSUF-Swin   & 15.52 & 0.4255 & 0.13941 & 0.03307 \\
% SSUF-MAE    & \textbf{16.06} & \textbf{0.4501} & 0.12924 & 0.03009 \\
% \hline
% \end{tabular}%
% }
% \end{table}

% \begin{table}[ht]
% \centering
% \caption{Classification performance comparison of SSUF-ConvAE, SSUF-ViT, SSUF-Swin, and SSUF-MAE on the Touch-and-Go dataset under the 10\% visible patch setting. Best results are highlighted in bold.}
% \label{tab:classification_results}
% \resizebox{\columnwidth}{!}{%
% \begin{tabular}{lcccc}
% \hline
% \textbf{Model} & \textbf{Accuracy $\uparrow$} & \textbf{Precision $\uparrow$} & \textbf{Recall $\uparrow$} & \textbf{F1-Score $\uparrow$} \\
% \hline
% SSUF-ConvAE & 83.90 & 0.8405 & 0.8390 & 0.8385 \\
% SSUF-ViT    & 86.30 & 0.8640 & 0.8630 & 0.8630 \\
% SSUF-Swin   & \textbf{89.21} & \textbf{0.8930} & \textbf{0.8921} & \textbf{0.8922} \\
% SSUF-MAE    & 76.71 & 0.7650 & 0.7671 & 0.7651 \\
% \hline
% \end{tabular}
% }
% \end{table}

\begin{table}[t]
\centering
\caption{Class-wise classification performance of the evaluated SSUF architectures on the Touch-and-Go test set under the 10\% visible patch setting. Precision, recall, and F1-score are reported separately for each material category.}
\label{tab:classwise_results}
\renewcommand{\arraystretch}{1.1}
\setlength{\tabcolsep}{8pt}
\begin{tabular}{llcccc}
\hline
\textbf{Model} & \textbf{Class} & \textbf{Precision} & \textbf{Recall} & \textbf{F1-Score} \\
\hline

\multirow{4}{*}{SSUF-ConvAE}
& Grass    & 0.9091 & 0.9375 & 0.9231  \\
& Concrete & 0.8596 & 0.7656 & 0.8099 \\
& Wood     & 0.8390 & 0.8866 & 0.8622 \\
& Rock     & 0.7431 & 0.7941 & 0.7678 \\
\hline

\multirow{4}{*}{SSUF-ViT}
& Grass    & 0.9381 & 0.9479 & 0.9430 \\
& Concrete & 0.8715 & 0.8125 & 0.8410 \\
& Wood     & 0.8593 & 0.8814 & 0.8702 \\
& Rock     & 0.7890 & 0.8431 & 0.8152 \\
\hline

\multirow{4}{*}{SSUF-Swin}
& Grass    & \textbf{0.9892} & \textbf{0.9583} & \textbf{0.9735} \\
& Concrete & \textbf{0.8883} & \textbf{0.8698} & \textbf{0.8789} \\
& Wood     & \textbf{0.8634} & \textbf{0.9124} & \textbf{0.8872}  \\
& Rock     & \textbf{0.8673} & \textbf{0.8333} & \textbf{0.8500}  \\
\hline

\multirow{4}{*}{SSUF-MAE}
& Grass    & 0.8426 & 0.9479 & 0.8922 \\
& Concrete & 0.7571 & 0.6979 & 0.7263 \\
& Wood     & 0.7739 & 0.7938 & 0.7837 \\
& Rock     & 0.6900 & 0.6765 & 0.6832 \\
\hline
\end{tabular}
\end{table}

\begin{table}[ht]
\centering
\caption{Model complexity comparison of the evaluated SSUF architectures.
Lower values indicate lower model complexity.}
\label{tab:model_complexity}

\footnotesize
\renewcommand{\arraystretch}{1.3}

\begin{tabular*}{\columnwidth}
{@{\extracolsep{\fill}}lccc@{}}
\hline
\rule{0pt}{4.2ex}

\textbf{Model} &
\shortstack{\textbf{Total}\\\textbf{Parameters}} &
\shortstack{\textbf{Trainable}\\\textbf{Parameters}} &
\shortstack{\textbf{Non-Trainable}\\\textbf{Parameters}} \\
\hline
SSUF-ConvAE & 38,923,847  & 15,336,711 & 23,587,136 \\
SSUF-ViT    & 92,028,039  & 5,635,335  & 86,392,704 \\
SSUF-Swin   & \textbf{32,300,049} & \textbf{4,718,375} &
\textbf{27,581,674} \\
SSUF-MAE    & 109,822,471 & 24,020,295 & 85,802,176 \\
\hline
\end{tabular*}

\end{table}

\begin{table}[ht]
\centering
\footnotesize\caption{Inference time comparison of SSUF-ConvAE, SSUF-ViT, SSUF-Swin, and SSUF-MAE on the Touch-and-Go test set. }
\label{tab:inference_time}
\resizebox{\columnwidth}{!}{%
\begin{tabular}{lccc}
\hline
\textbf{Model} &
\makecell{\textbf{Total Time}\\\textbf{(s) $\downarrow$}} &
\makecell{\textbf{Avg. Time/Img}\\\textbf{(s) $\downarrow$}} &
\makecell{\textbf{Avg. Time/Img}\\\textbf{(ms) $\downarrow$}} \\
\hline
SSUF-ConvAE & 2.329 & 0.003988 & 3.988 \\
SSUF-ViT    & 2.784 & 0.004766 & 4.766 \\
SSUF-Swin   & 2.257 & 0.003865 & 3.865 \\
SSUF-MAE    & \textbf{1.481} & \textbf{0.002536} & \textbf{2.536} \\
\hline
\end{tabular}
}
\end{table}

\begin{figure*}[!t]
\centering

%================ ROC CURVES ====================

\begin{minipage}{0.23\textwidth}
    \centering
    \includegraphics[width=\linewidth]{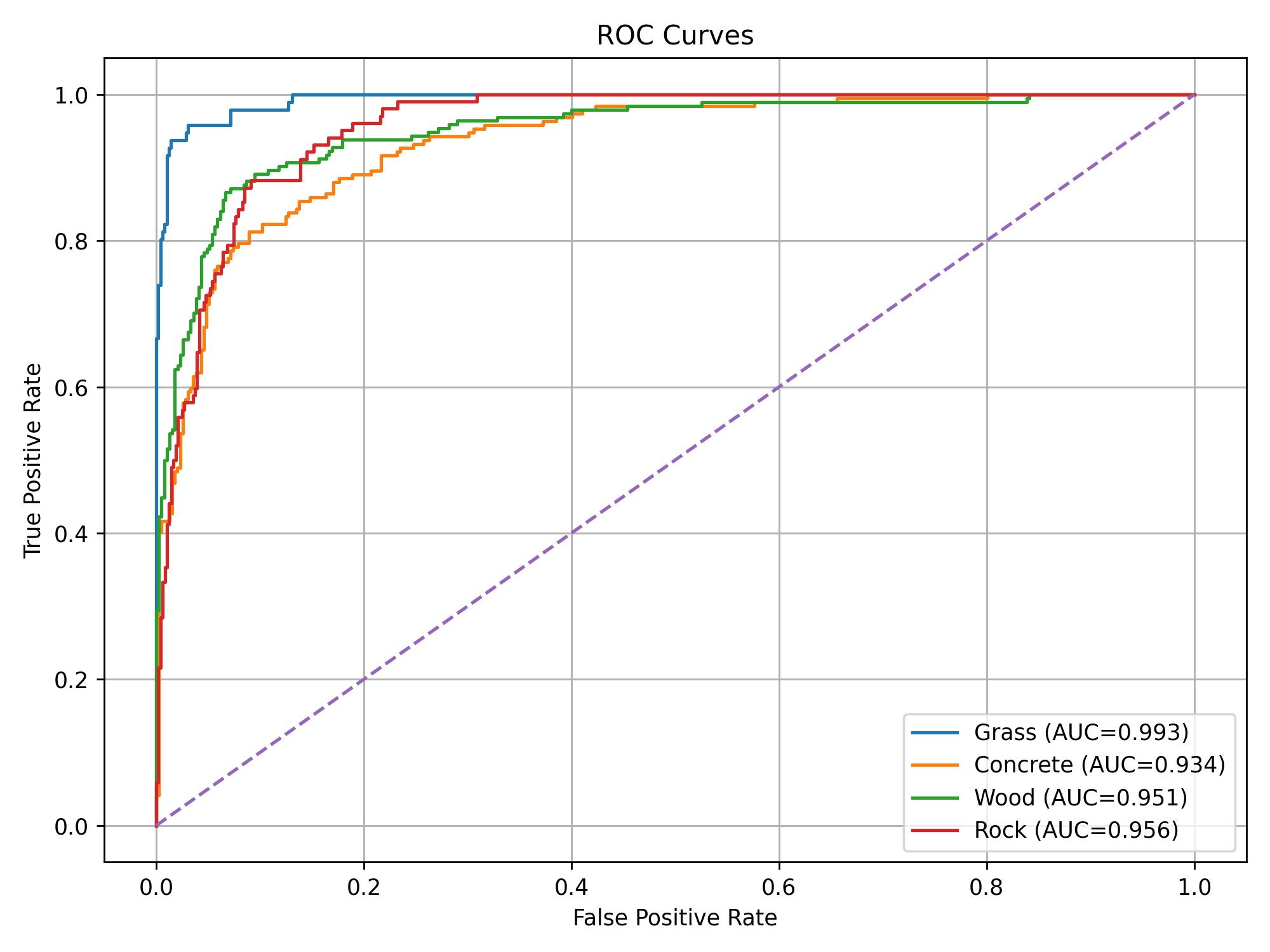}\\[-1mm]
    {\footnotesize SSUF-ConvAE}
\end{minipage}
\hfill
\begin{minipage}{0.23\textwidth}
    \centering
    \includegraphics[width=\linewidth]{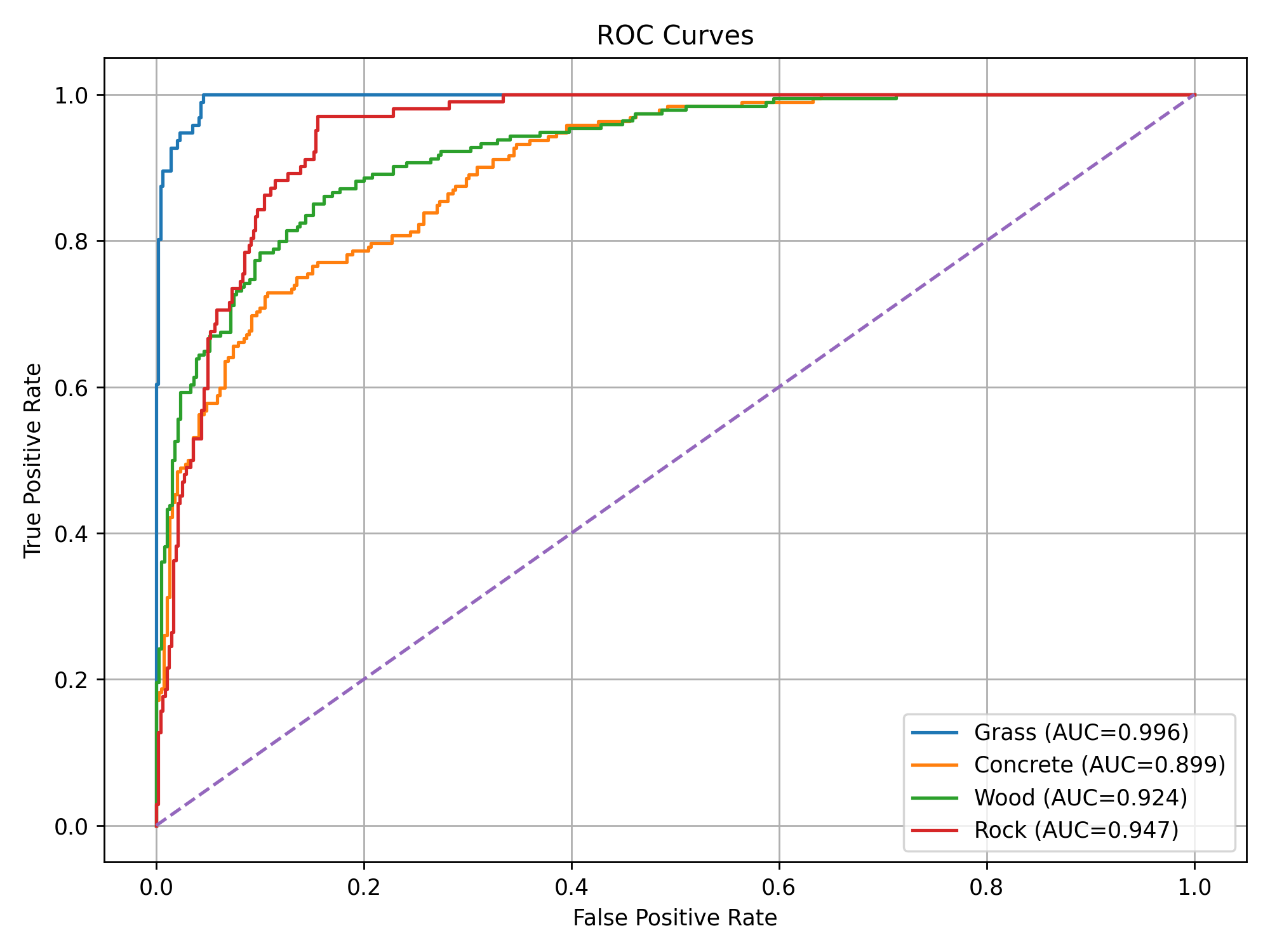}\\[-1mm]
    {\footnotesize SSUF-MAE}
\end{minipage}
\hfill
\begin{minipage}{0.23\textwidth}
    \centering
    \includegraphics[width=\linewidth]{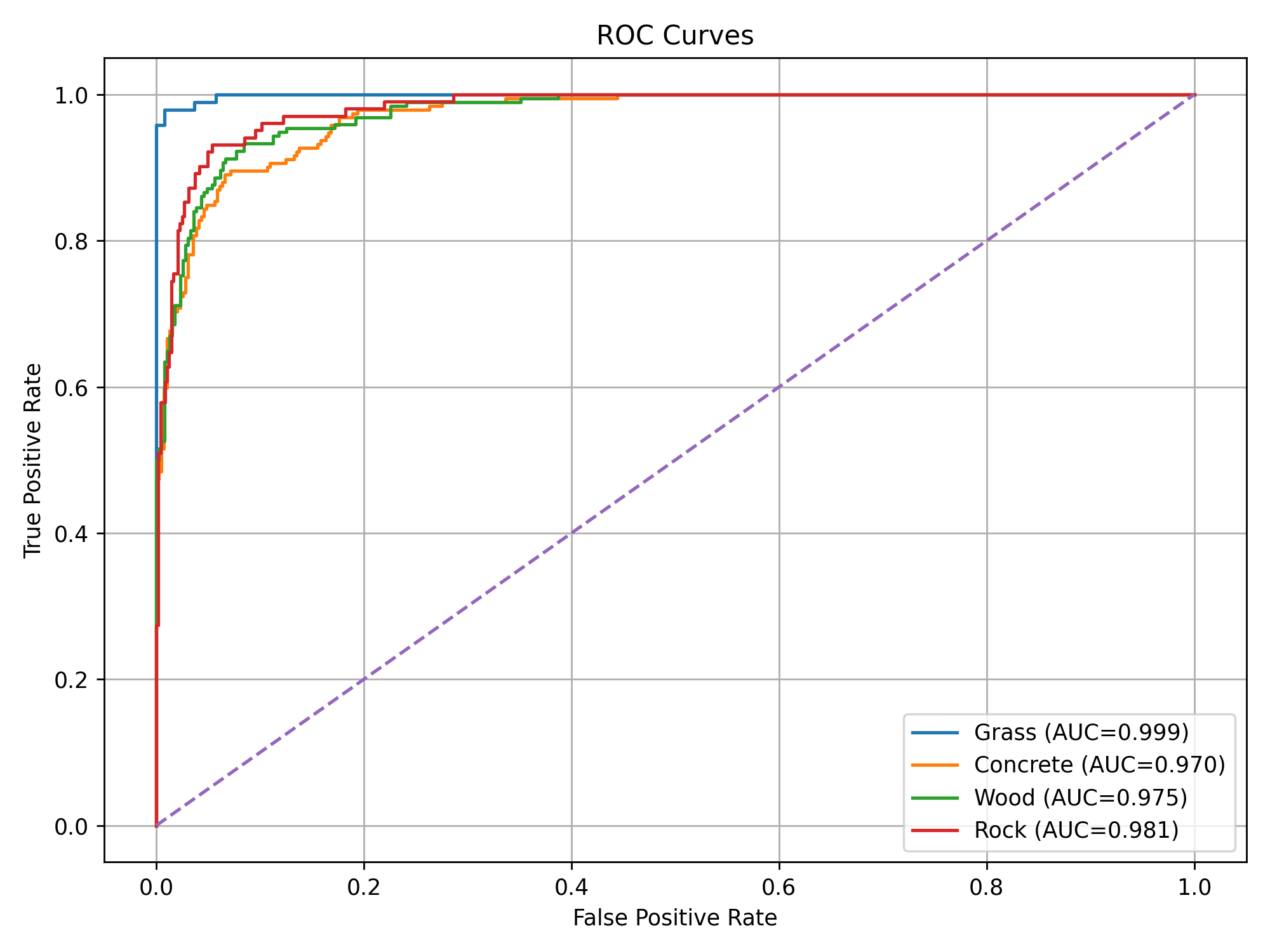}\\[-1mm]
    {\footnotesize SSUF-Swin}
\end{minipage}
\hfill
\begin{minipage}{0.23\textwidth}
    \centering
    \includegraphics[width=\linewidth]{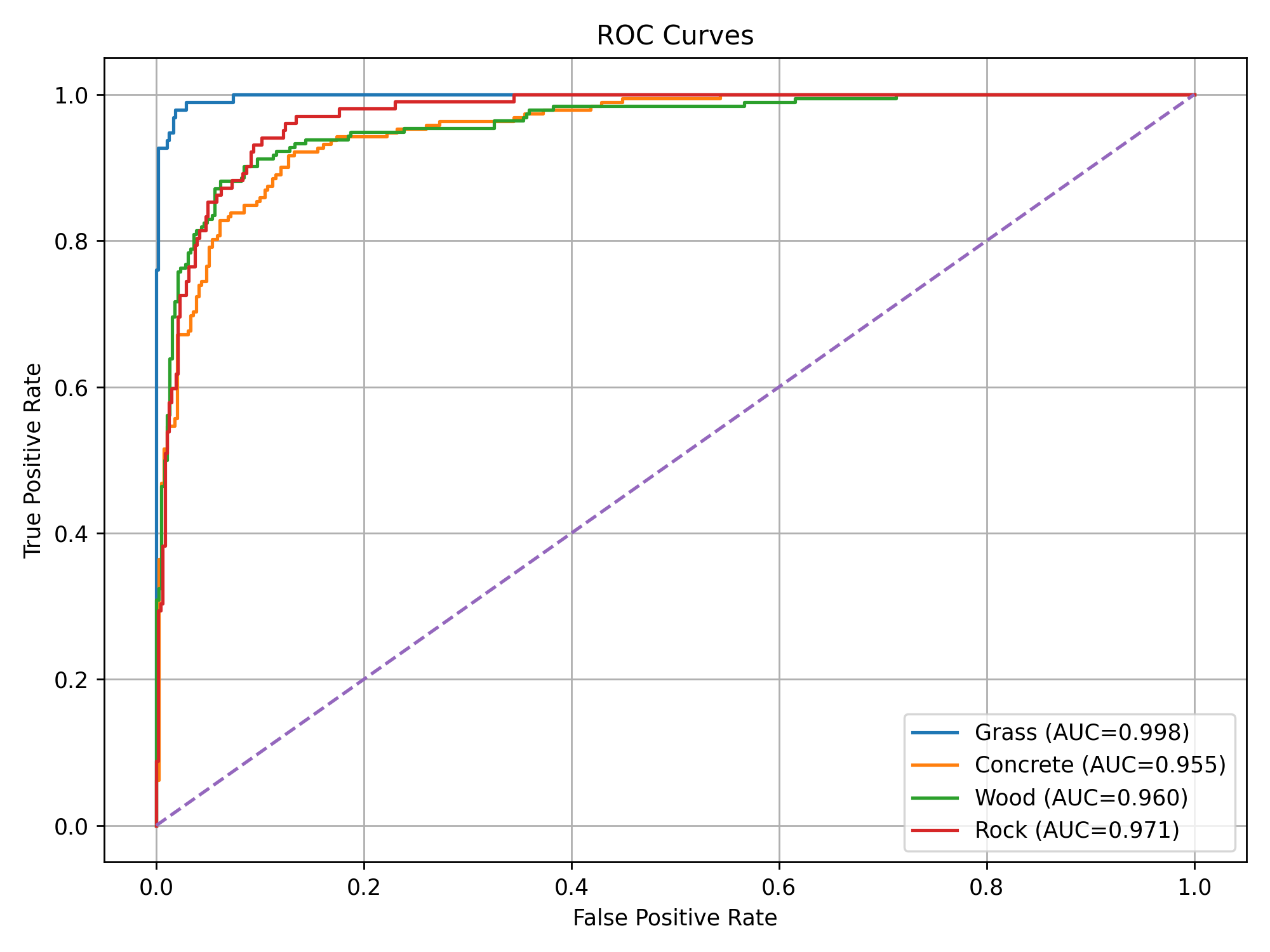}\\[-1mm]
    {\footnotesize SSUF-ViT}
\end{minipage}

\vspace{3mm}

{\footnotesize (a) ROC Curves}

\vspace{4mm}

%================ CONFUSION MATRICES ====================

\begin{minipage}{0.23\textwidth}
    \centering
    \includegraphics[width=\linewidth]{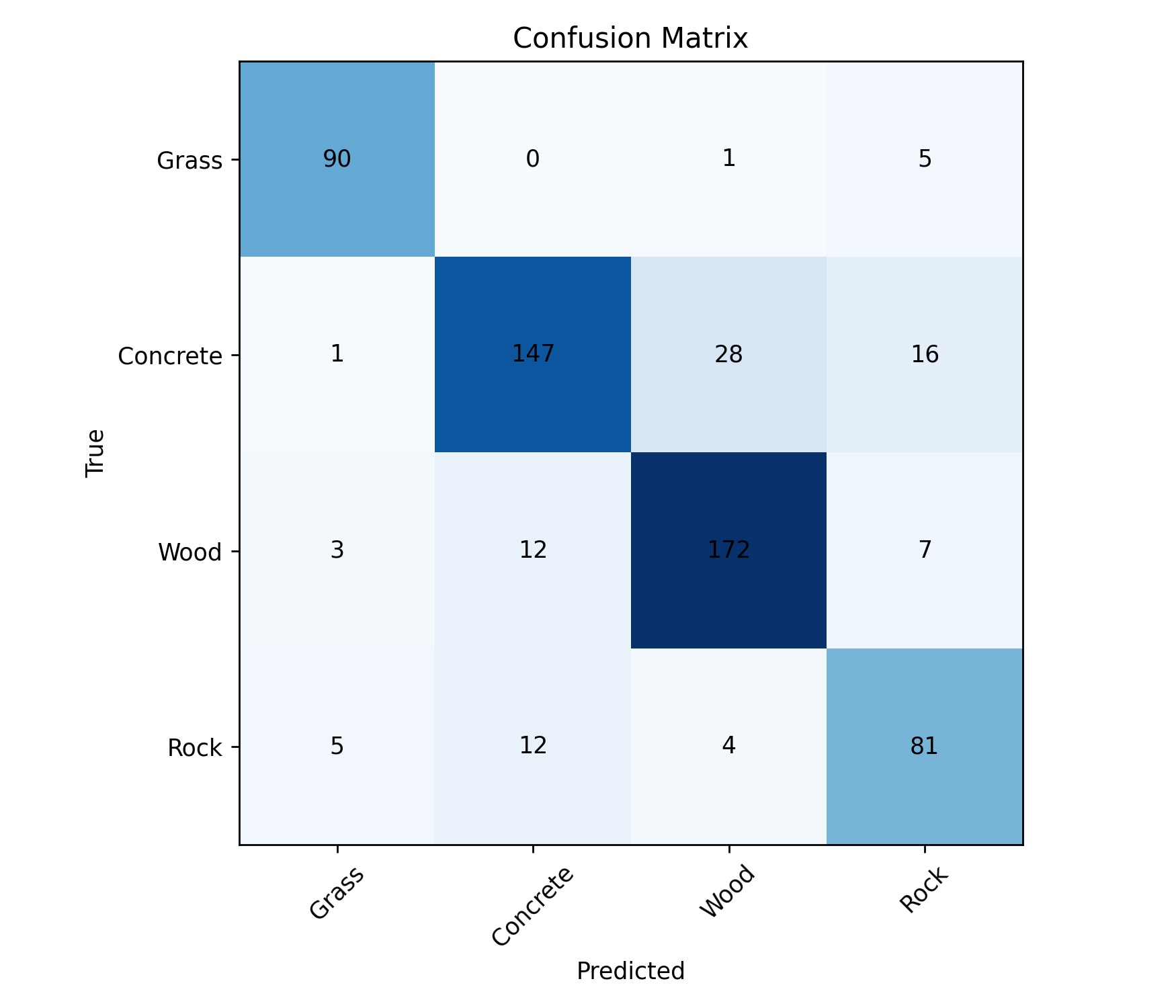}\\[-1mm]
    {\footnotesize SSUF-ConvAE}
\end{minipage}
\hfill
\begin{minipage}{0.23\textwidth}
    \centering
    \includegraphics[width=\linewidth]{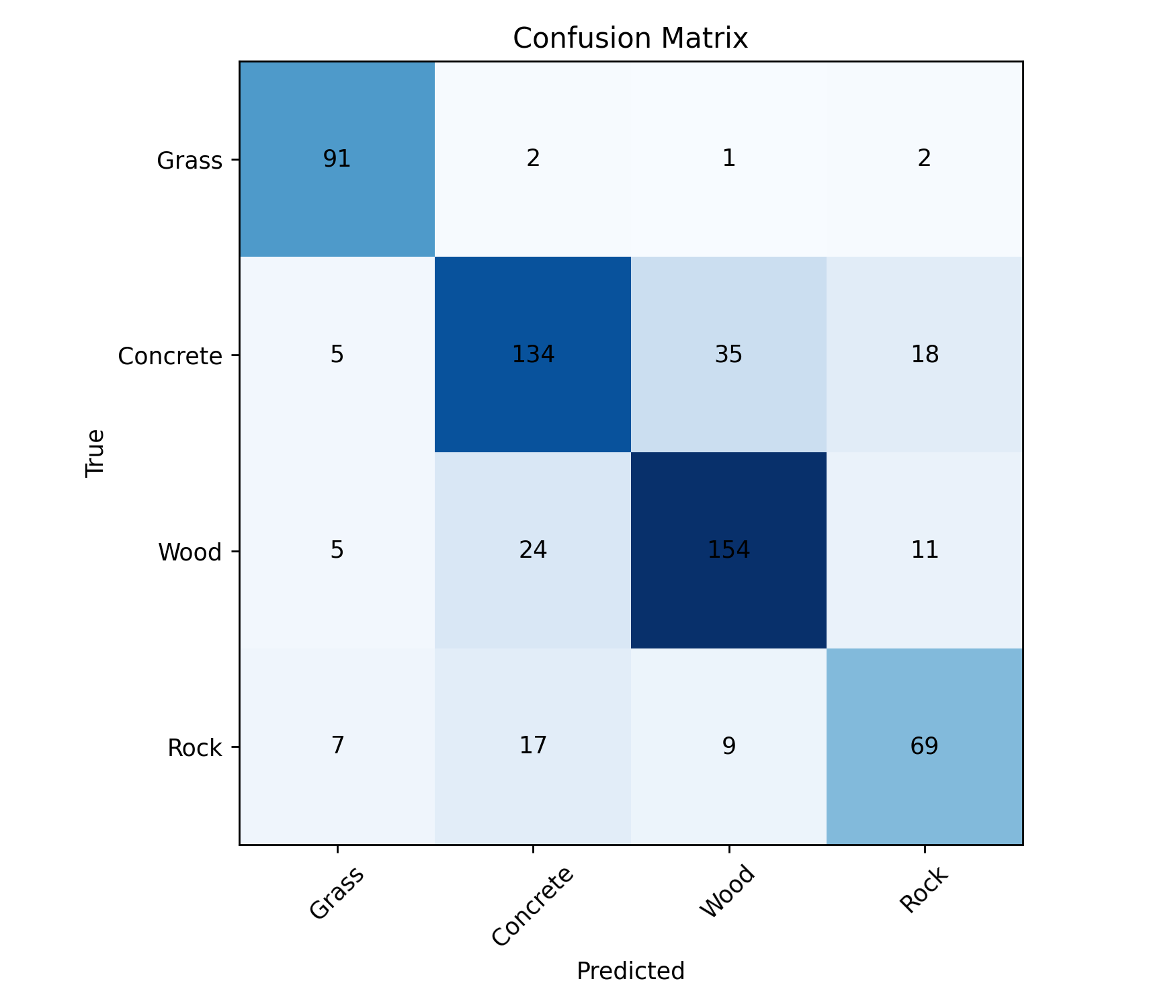}\\[-1mm]
    {\footnotesize SSUF-MAE}
\end{minipage}
\hfill
\begin{minipage}{0.23\textwidth}
    \centering
    \includegraphics[width=\linewidth]{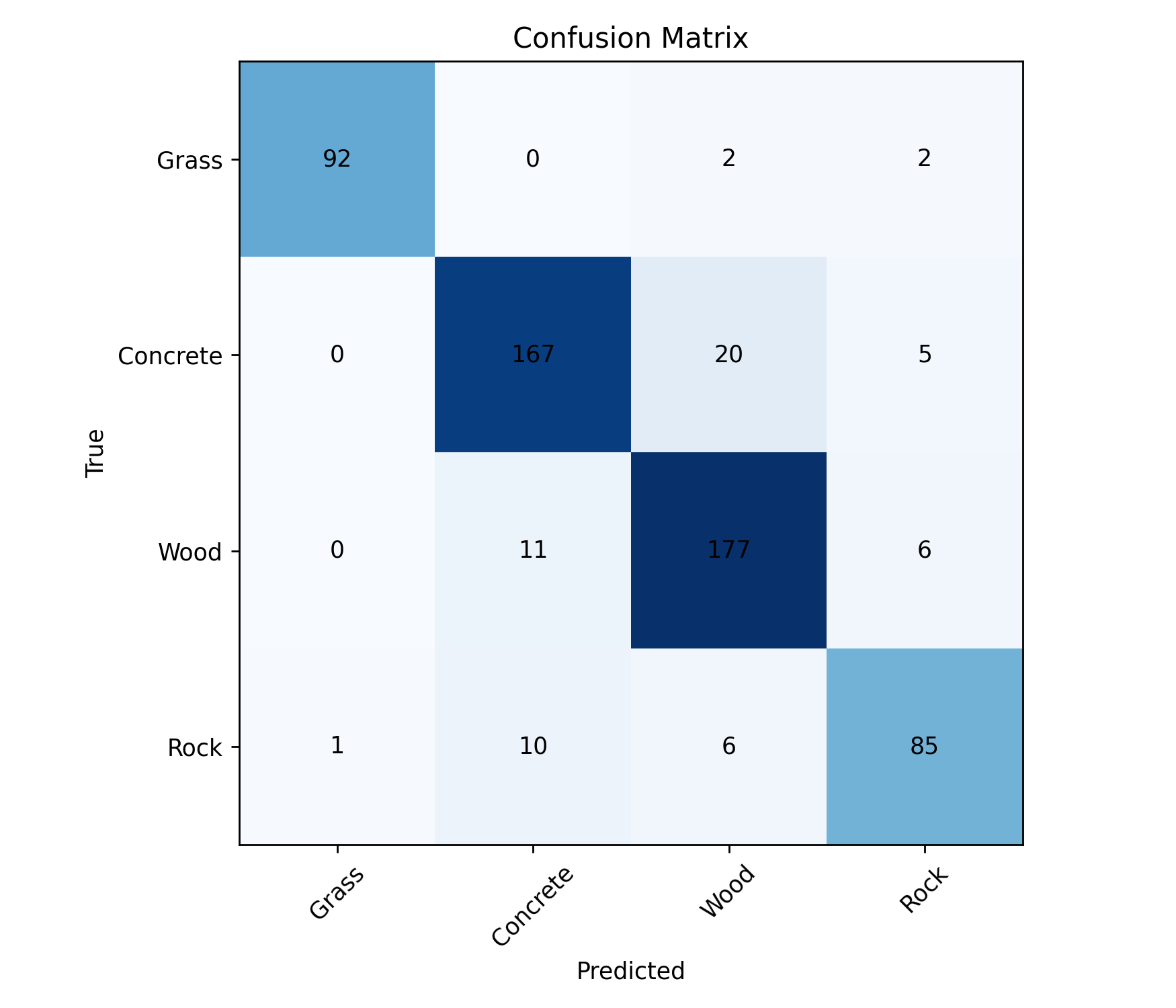}\\[-1mm]
    {\footnotesize SSUF-Swin}
\end{minipage}
\hfill
\begin{minipage}{0.23\textwidth}
    \centering
    \includegraphics[width=\linewidth]{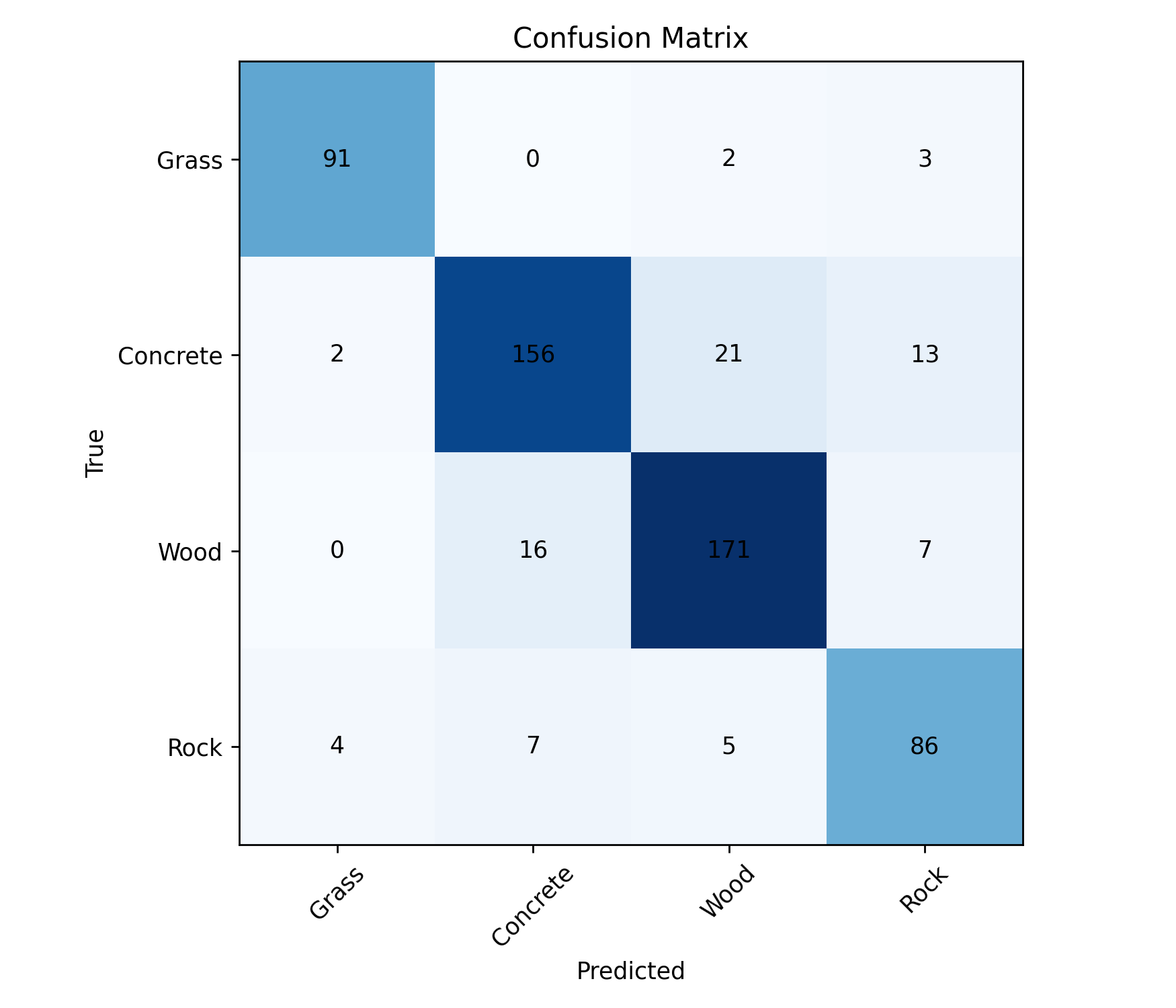}\\[-1mm]
    {\footnotesize SSUF-ViT}
\end{minipage}

\vspace{3mm}

{\footnotesize (b) Confusion Matrices}

\vspace{-1mm}

\caption{ROC curves (top row) and confusion matrices (bottom row) for
the evaluated SSUF architectures on the Touch-and-Go dataset under the
10\% visible-patch setting. SSUF-Swin achieved the strongest overall
classification performance, followed by SSUF-ViT, SSUF-ConvAE, and
SSUF-MAE.}
\label{fig:roc_cm_combined}

\end{figure*}

\subsection{Dataset Description}

Experiments were conducted on the Touch-and-Go dataset \cite{yang2022touch}, a large-scale outdoor surface perception dataset containing images collected from real-world interactions with diverse ground materials. The dataset includes four surface categories: Grass, Concrete, Wood, and Rock, representing a range of textures and visual characteristics commonly encountered in outdoor environments.

All images were resized to $224 \times 224$ pixels and normalized according to the requirements of the corresponding pretrained backbone. Following the protocol described in Section III, sparse observations were generated by retaining only a random continuous patch corresponding to 10\% of the original image area. The dataset contains a total of 2,921 images, comprising 416 Grass, 1,000 Concrete, 1,000 Wood, and 505 Rock samples. Following the 60:20:20 partitioning strategy, 1,753 images were used for training, while 584 images were allocated to both the validation and test sets.

% The dataset was partitioned into training, validation, and test sets using a 60:20:20 split to ensure unbiased performance evaluation. 

\subsection{Image Reconstruction Performance}

% Table~\ref{tab:recon_metrics} presents a quantitative comparison of reconstruction performance under the random continuous patch visibility constraint. The proposed Partial Convolution (PConv) approach outperforms all baseline architectures across every metric, achieving a peak signal-to-noise ratio (PSNR) of 17.33 dB and a structural similarity index (SSIM) of 0.5694. Furthermore, it minimizes reconstruction error, yielding the lowest observed MSE (0.02337) and MAE (0.1029).

Table~\ref{tab:reconstruction_results} summarizes the reconstruction performance of the evaluated SSUF architectures under the 10\% visible patch setting. SSUF-MAE achieved the highest PSNR (16.06 dB) and SSIM (0.4501), among the evaluated models. SSUF-ViT produced the moderate reconstruction errors, obtaining the best MAE (0.12861) and MSE (0.02940). SSUF-ConvAE also delivered comparable reconstruction results, achieving performance close to that of SSUF-MAE and SSUF-ViT despite its comparatively simpler convolutional design. In contrast, SSUF-Swin exhibited lower reconstruction quality across all evaluated metrics, suggesting that its hierarchical transformer representations are better suited for discriminative recognition tasks than pixel-level image recovery. Overall, the results indicate that SSUF-MAE is the most effective architecture for reconstruction.

\subsection{Material Classification Performance}

Table~\ref{tab:classification_results} summarizes the material classification performance of the evaluated SSUF architectures. SSUF-Swin achieved the best overall results, obtaining the highest accuracy (89.21\%), precision (0.8930), recall (0.8921), and F1-score (0.8922), demonstrating its superior ability to discriminate surface materials from highly sparse visual observations. SSUF-ViT delivered the second-best performance with an accuracy of 86.30\% and an F1-score of 0.8630, providing a strong balance between reconstruction and classification tasks. SSUF-ConvAE achieved competitive performance with an accuracy of 83.90\%, while SSUF-MAE obtained the lowest classification accuracy (76.71\%) despite its strong reconstruction capability. Overall, the results indicate that transformer-based architectures are more effective for sparse surface material recognition, with the hierarchical feature representations and shifted-window attention mechanism of SSUF-Swin contributing to its superior classification performance.

To further examine the overall classification results, Table~\ref{tab:classwise_results} presents class-wise precision, recall, and F1-score for the four material categories. Although Concrete and Wood contain more samples than Grass and Rock, balanced class weights and data augmentation were used during training to reduce the effect of class imbalance. Grass achieved the highest F1-score across all models, while Rock was more challenging for SSUF-ConvAE and SSUF-MAE. SSUF-Swin showed the most consistent class-wise performance, with
F1-scores ranging from 0.8500 to 0.9735.

\subsection{Computational Complexity Comparison}

Table~\ref{tab:model_complexity} compares the parameter complexity of the evaluated SSUF architectures. SSUF-MAE is the largest model, containing 109.8 million parameters, including 24.0 million trainable parameters, reflecting its substantial representational capacity. SSUF-ViT follows with 92.0 million total parameters but requires only 5.6 million trainable parameters due to extensive transfer learning. In contrast, SSUF-Swin is the most parameter-efficient architecture, containing only 32.3 million parameters and 4.7 million trainable parameters while achieving the highest classification performance. SSUF-ConvAE contains 38.9 million parameters, positioning it between the transformer-based and reconstruction-oriented models. These results indicate that SSUF-Swin provides an effective balance between model complexity and predictive performance, whereas SSUF-MAE and SSUF-ViT leverage larger pretrained representations to improve reconstruction quality under sparse visual observations.

\subsection{Inference Efficiency Comparison}

Table~\ref{tab:inference_time} presents the inference efficiency of the evaluated SSUF architectures on the Touch-and-Go test set. SSUF-MAE achieved the fastest inference speed, requiring 1.481 seconds to process the entire test set and an average of 2.536 ms per image. SSUF-Swin exhibited the second-best efficiency with an average inference time of 3.865 ms per image, followed closely by SSUF-ConvAE at 3.988 ms per image. SSUF-ViT showed the highest inference latency, requiring 4.766 ms per image. Despite these differences, all architectures achieved real-time performance with average processing times below 5 ms per image. Combined with its superior classification accuracy, SSUF-Swin provides an attractive balance between predictive performance and computational efficiency, whereas SSUF-MAE offers the fastest inference and strongest reconstruction capability.

% \begin{figure*}[t]
% \centering

% \begin{subfigure}{0.4\textwidth}
%     \centering
%     \includegraphics[width=\linewidth]{roc-cm/cae_roc_curves.png}
%     \caption{Convolutional Autoencoder}
%     \label{fig:roc_cae}
% \end{subfigure}
% \hfill
% \begin{subfigure}{0.4\textwidth}
%     \centering
%     \includegraphics[width=\linewidth]{roc-cm/mae_roc_curves.png}
%     \caption{Masked Autoencoder (MAE)}
%     \label{fig:roc_mae}
% \end{subfigure}

% \vspace{0.2cm}

% \begin{subfigure}{0.4\textwidth}
%     \centering
%     \includegraphics[width=\linewidth]{roc-cm/swin_roc_curves.png}
%     \caption{Swin Transformer}
%     \label{fig:roc_swin}
% \end{subfigure}
% \hfill
% \begin{subfigure}{0.4\textwidth}
%     \centering
%     \includegraphics[width=\linewidth]{roc-cm/vit_roc_curves.png}
%     \caption{Vision Transformer (ViT)}
%     \label{fig:roc_vit}
% \end{subfigure}

% \caption{Receiver Operating Characteristic (ROC) curves for the four evaluated models on the Touch-and-Go dataset under the 10\% visible patch setting. Each subplot presents the one-vs-rest ROC performance for Grass, Concrete, Wood, and Rock classes. Swin Transformer achieved the highest overall class separability, followed by ViT, ConvAE, and MAE.}
% \label{fig:roc_comparison}
% \end{figure*}

\subsection{ROC Curves of Models}

Fig. ~\ref{fig:roc_cm_combined}(a) presents the ROC curves of the four evaluated SSUF architectures for multi-class surface material classification on the Touch-and-Go dataset. Across all architectures, the Grass class consistently achieved the highest AUC values, ranging from 0.995 to 0.999, indicating that it is the most distinguishable material category even when only a small portion of the image is available. Swin Transformer exhibited the strongest overall class discrimination performance, achieving AUC values of 0.999, 0.970, 0.975, and 0.981 for Grass, Concrete, Wood, and Rock, respectively. ViT also demonstrated strong separability across all classes, with AUC values exceeding 0.95. ConvAE achieved competitive performance, particularly for Grass (0.993) and Rock (0.956), while MAE showed comparatively lower AUC values for Concrete (0.899) and Wood (0.924). Overall, the ROC analysis confirms the classification results presented in Table~\ref{tab:classification_results}, demonstrating that transformer-based architectures, especially Swin Transformer, provide superior robustness and class separability for surface material recognition from highly sparse visual observations.

% Figure \ref{fig:roc_curves} shows the one-vs-rest ROC curves of all six models on the Touch and Go dataset for the four terrain classes: Concrete, Grass, Wood, and Rock. Across all models, the Grass class shows the strongest separability, with AUC values close to or equal to 1.0. The Conv Autoencoder produces the most consistent ROC performance across all classes, with high AUC values for Concrete, Wood, and Rock, and perfect separation for Grass. SMARC/PConv also performs strongly, showing balanced ROC curves across all four classes, especially for Concrete, Wood, and Rock.

% Comparing the models, Conv Autoencoder shows the best overall ROC behavior, followed by SMARC/PConv. DETR performs reasonably well, especially for Grass and Rock, but its Concrete and Wood curves are lower than CAE and SMARC. ViT and Swin show moderate performance, with some class-wise variation, while MAE shows weaker separation for Rock and Wood compared with the other models. Overall, the ROC curves indicate that CAE and SMARC/PConv are more effective in distinguishing terrain classes under random continuous patch visibility conditions.

\subsection{Confusion Matrices of Models}

Fig. ~\ref{fig:roc_cm_combined}(b) illustrates the confusion matrices obtained by SSUF-ConvAE, MAE, Swin Transformer, and ViT under the 10\% visible patch setting. All models achieved high recognition rates for the Grass class, with 91-92 correctly classified samples. The primary source of classification error was observed between the Concrete and Wood categories, suggesting substantial visual similarity under severe information loss. Swin Transformer produced the most pronounced diagonal dominance, correctly classifying 92 Grass, 167 Concrete, 177 Wood, and 85 Rock samples while maintaining the fewest overall misclassifications. ViT achieved the second-best performance, correctly identifying 91 Grass, 156 Concrete, 171 Wood, and 86 Rock samples. Although all models achieved strong performance, ConvAE and MAE showed greater confusion between Concrete and Wood samples. Swin Transformer produced the fewest misclassifications overall, followed by ViT, which is consistent with their higher classification accuracies reported in Table~\ref{tab:classification_results}. These results indicate that transformer-based models are more effective for surface material recognition under highly sparse visual observations.

%{\color{blue}
%\subsection{Limitations and Future Work}

%There are a few limitations remain in the current study. First, classification errors are more frequent between Concrete and Wood, suggesting that visually similar surface categories remain challenging under extreme visual sparsity. Future work will investigate more discriminative representations for these closely related materials.

%Second, the present study compares only the proposed dual-task architectures and does not include corresponding single-task baselines trained under the same 10\% visible-patch setting. Future work will compare the dual-task models with single-task reconstruction and classification counterparts to better quantify the effect of joint learning.

%Finally, although equal weights ($\lambda_{\mathrm{rec}} = \lambda_{\mathrm{cls}} = 1.0$) were used in the current experiments, the reconstruction and classification losses operate on different numerical scales. A systematic ablation of the loss weights was not performed and will be investigated in future work to determine a more effective balance between the two objectives.
%}

\section{Conclusion}

% This paper presented the Sparse Surface Understanding Framework (SSUF), a unified dual-task framework for simultaneous surface reconstruction and material classification from highly sparse visual observations. Four pretrained architectures-ConvAE, ViT, Swin Transformer, and MAE were adapted to perform both tasks and evaluated using only a random continuous patch corresponding to 10\% of the original image area from the Touch-and-Go dataset. Experimental results revealed distinct strengths among the evaluated models. SSUF-Swin achieved the best classification performance with an accuracy of 89.21\%, while SSUF-MAE produced the highest reconstruction quality with a PSNR of 16.06 dB and an SSIM of 0.4501. SSUF-ViT provided the most balanced performance across both reconstruction and classification tasks. Overall, the results demonstrate that meaningful surface characteristics can be reconstructed and recognized even under extreme visual sparsity, highlighting the effectiveness of pretrained transformer representations for sparse surface understanding. Future work will explore adaptive masking strategies, multimodal visual-tactile fusion, and self-supervised learning approaches to further improve robustness and generalization in real-world robotic perception environments. 
This paper presented the Sparse Surface Understanding Framework (SSUF), a unified dual-task framework for simultaneous surface reconstruction and material classification from highly sparse visual observations. Four pretrained architectures-ConvAE, ViT, Swin Transformer, and MAE-were adapted to perform both tasks and evaluated using a random continuous patch corresponding to approximately 10\% of the original image area from the Touch-and-Go dataset. Experimental results showed different strengths across the evaluated models. SSUF-Swin achieved the best classification performance with an accuracy of 89.21\%, while SSUF-MAE obtained the highest reconstruction scores with a PSNR of 16.06 dB and an SSIM of 0.4501. SSUF-ViT provided a competitive balance between reconstruction and classification performance. The results show that useful material cues can still be learned from highly sparse visual inputs, particularly for classification. Reconstruction remains more difficult under this setting, with the models mainly recovering coarse appearance rather than fine visual details. These observations highlight the different strengths of the evaluated architectures under extreme visual sparsity. Future work will explore adaptive masking strategies, multimodal visual-tactile fusion, and self-supervised learning approaches to further improve robustness and generalization in real-world robotic perception environments.

\section*{Acknowledgment}
The authors acknowledge the support and resources provided by the Bioinspired Robotics, AI, Imaging and Neurocognitive Systems (BRAINS) Laboratory at The University of Alabama.

\bibliographystyle{IEEEtran}
\bibliography{references}

@article{dosovitskiy2020image,
  title={An image is worth 16x16 words: Transformers for image recognition at scale},
  author={Dosovitskiy, Alexey and Beyer, Lucas and Kolesnikov, Alexander and Weissenborn, Dirk and Zhai, Xiaohua and Unterthiner, Thomas and Dehghani, Mostafa and Minderer, Matthias and Heigold, Georg and Gelly, Sylvain and others},
  journal={arXiv preprint arXiv:2010.11929},
  year={2020}
}

@inproceedings{liu2022swin,
  title={Swin transformer v2: Scaling up capacity and resolution},
  author={Liu, Ze and Hu, Han and Lin, Yutong and Yao, Zhuliang and Xie, Zhenda and Wei, Yixuan and Ning, Jia and Cao, Yue and Zhang, Zheng and Dong, Li and others},
  booktitle={Proceedings of the IEEE/CVF conference on computer vision and pattern recognition},
  pages={12009--12019},
  year={2022}
}

@inproceedings{he2022masked,
  title={Masked autoencoders are scalable vision learners},
  author={He, Kaiming and Chen, Xinlei and Xie, Saining and Li, Yanghao and Doll{\'a}r, Piotr and Girshick, Ross},
  booktitle={Proceedings of the IEEE/CVF conference on computer vision and pattern recognition},
  pages={16000--16009},
  year={2022}
}

@inproceedings{he2016deep,
  title={Deep residual learning for image recognition},
  author={He, Kaiming and Zhang, Xiangyu and Ren, Shaoqing and Sun, Jian},
  booktitle={Proceedings of the IEEE conference on computer vision and pattern recognition},
  pages={770--778},
  year={2016}
}

@article{yang2022touch,
  title={Touch and go: Learning from human-collected vision and touch},
  author={Yang, Fengyu and Ma, Chenyang and Zhang, Jiacheng and Zhu, Jing and Yuan, Wenzhen and Owens, Andrew},
  journal={arXiv preprint arXiv:2211.12498},
  year={2022}
}

@incollection{koonce2021resnet,
  title={ResNet 50},
  author={Koonce, Brett},
  booktitle={Convolutional neural networks with swift for tensorflow: image recognition and dataset categorization},
  pages={63--72},
  year={2021},
  publisher={Springer}
}

@inproceedings{bell2015material,
  title={Material recognition in the wild with the materials in context database},
  author={Bell, Sean and Upchurch, Paul and Snavely, Noah and Bala, Kavita},
  booktitle={Proceedings of the IEEE conference on computer vision and pattern recognition},
  pages={3479--3487},
  year={2015}
}

@inproceedings{pathak2016context,
  title={Context encoders: Feature learning by inpainting},
  author={Pathak, Deepak and Krahenbuhl, Philipp and Donahue, Jeff and Darrell, Trevor and Efros, Alexei A},
  booktitle={Proceedings of the IEEE conference on computer vision and pattern recognition},
  pages={2536--2544},
  year={2016}
}

@inproceedings{liu2018image,
  title={Image inpainting for irregular holes using partial convolutions},
  author={Liu, Guilin and Reda, Fitsum A and Shih, Kevin J and Wang, Ting-Chun and Tao, Andrew and Catanzaro, Bryan},
  booktitle={Proceedings of the European conference on computer vision (ECCV)},
  pages={85--100},
  year={2018}
}

@article{yang2024review,
  title={Review of deep learning-based image inpainting techniques},
  author={Yang, Jing and Ruhaiyem, Nur Intan Raihana},
  journal={IEEE Access},
  volume={12},
  pages={138441--138482},
  year={2024},
  publisher={IEEE}
}

@article{penchala2026one,
  title={One Patch Is All You Need: Joint Surface Material Reconstruction and Classification from Minimal Visual Cues},
  author={Penchala, Sindhuja and Money, Gavin and Marques, Gabriel and Wood, Samuel and Kirschman, Jessica and Atkison, Travis and Rahimi, Shahram and Amiri Golilarz, Noorbakhsh},
  journal={Sensors},
  volume={26},
  number={7},
  pages={2083},
  year={2026},
  publisher={MDPI}
}

@article{kansana2025surformer,
  title={Surformer v2: A multimodal classifier for surface understanding from touch and vision},
  author={Kansana, Manish and Penchala, Sindhuja and Rahimi, Shahram and Golilarz, Noorbakhsh Amiri},
  journal={arXiv preprint arXiv:2509.04658},
  year={2025}
}

@article{kansana2025surformerv1,
  title={Surformer v1: Transformer-based surface classification using tactile and vision features},
  author={Kansana, Manish and Hossain, Elias and Rahimi, Shahram and Amiri Golilarz, Noorbakhsh},
  journal={Information},
  volume={16},
  number={10},
  pages={839},
  year={2025},
  publisher={MDPI}
}

@article{golilarz2026towards,
  title={Towards neurocognitive-inspired intelligence: From ai’s structural mimicry to human-like functional cognition},
  author={Golilarz, Noorbakhsh Amiri and Al Khatib, Hassan S and Rahimi, Shahram},
  journal={IEEE Access},
  year={2026},
  publisher={IEEE}
}

@article{golilarz2025bridging,
  title={Bridging the gap: Toward cognitive autonomy in artificial intelligence},
  author={Golilarz, Noorbakhsh Amiri and Penchala, Sindhuja and Rahimi, Shahram},
  journal={arXiv preprint arXiv:2512.02280},
  year={2025}
}

\end{document}